\documentclass[conference]{IEEEtran}
\IEEEoverridecommandlockouts

\usepackage{cite}
\usepackage{amsmath,amssymb,amsfonts}
\usepackage{algorithmic}
\usepackage{graphicx}
\usepackage{subcaption}
\usepackage{textcomp}
\usepackage{xcolor}
\usepackage[hyphens]{url}
\usepackage{listings}
\usepackage{tabularx}
\usepackage{booktabs}
\usepackage{multirow}
\usepackage{multicol}
\usepackage{adjustbox}
\usepackage{makecell}
\usepackage{xspace}
\usepackage{stfloats}
\usepackage{ragged2e} 
\usepackage{microtype} 

\lstdefinestyle{cppstyle}{
    language=C++,
    backgroundcolor=\color{black!5},
    commentstyle=\color{green!60!black},
    keywordstyle=\color{blue},
    numberstyle=\tiny\color{gray},
    stringstyle=\color{purple},
    basicstyle=\ttfamily\footnotesize,
    breakatwhitespace=false,
    breaklines=true,
    captionpos=b,
    keepspaces=true,
    numbers=none,
    numbersep=5pt,
    showspaces=false,
    showstringspaces=false,
    showtabs=false,
    tabsize=2
}
\lstdefinelanguage{codediff}{
  morecomment=[f][\color{red!80!black}]{-},
  morecomment=[f][\color{green!60!black}]{+},
  basicstyle=\ttfamily,
  commentstyle=\color{gray},
}

\newcolumntype{Y}{>{\centering\arraybackslash}X}
\newcolumntype{L}{>{\RaggedRight\arraybackslash}X}

\def\BibTeX{{\rm B\kern-.05em{\sc i\kern-.025em b}\kern-.08em
    T\kern-.1667em\lower.7ex\hbox{E}\kern-.125emX}}

\newcommand{\alphaarchitect}{ArchAgent\xspace}
\newcommand{\anonevolve}{AlphaEvolve\xspace}

\newcommand{\sisetup}[1]{}

\begin{document}

\pdfpagewidth=8.5in
\pdfpageheight=11in

\newcommand{\iscasubmissionnumber}{NaN}

\pagenumbering{arabic}

\title{ArchAgent v2: A Case Study with the Data Prefetching Championship}

\author{
  Abraham Gonzalez\textsuperscript{1}, 
  Raghav Gupta\textsuperscript{2$\dagger$}, 
  Akanksha Jain\textsuperscript{1}, 
  Hanna Alam\textsuperscript{1},
  Alexander Novikov\textsuperscript{3}, 
  Po-Sen Huang\textsuperscript{3}, \\[0.5ex]
  Matej Balog\textsuperscript{3}, 
  Marvin Eisenberger\textsuperscript{3}, 
  Sergey Shirobokov\textsuperscript{3}, 
  Ng{\^a}n V{\~u}\textsuperscript{3}, \\[0.5ex]
  Hank Levy\textsuperscript{1},
  Borivoje Nikoli{\'c}\textsuperscript{2}, 
  Sagar Karandikar\textsuperscript{2$\dagger$},
  Martin Dixon\textsuperscript{1}, 
  Parthasarathy Ranganathan\textsuperscript{1} \\[1ex]
  \textsuperscript{1}Google \qquad
  \textsuperscript{2}University of California, Berkeley \qquad
  \textsuperscript{3}Google DeepMind \\[0.5ex]
  \textit{\{abegonzalez, avjain, hannaalam, mgdixon, hanklevy, parthas\}@google.com} \\
  \textit{raghavgupta@berkeley.edu, \{bora, sagark\}@eecs.berkeley.edu} \\
  \textit{\{anovikov, posenhuang, matejb, meisenberger, shirobokov, nganvu\}@google.com}
  \thanks{\textsuperscript{$\dagger$}Work partially done while the author was affiliated with Google.}
}

\maketitle
\thispagestyle{plain}
\pagestyle{plain}

\begin{abstract}
Agentic artificial intelligence has shown great promise in automating algorithm design, but scaling similar techniques to computer microarchitecture discovery remains challenging due to vast search spaces, strict hardware budgets, and long simulation times.
In this work, we present \alphaarchitect v2, a framework which scales automated microarchitecture search to multi-level data prefetching.
While the original \alphaarchitect successfully discovered single-level cache replacement policies in competition settings, it does not scale to multi-level prefetching where the design space and degrees of freedom are larger.
To overcome this, we introduce two new additions to \alphaarchitect: a cascaded evolutionary search that subdivides the design space by sequentially evolving and freezing prefetchers at individual cache levels, and a hardware-realizability feedback loop that embeds real-time size-estimation directly into the evolution process. 

Evaluated under identical rules of the Fourth Data Prefetching Championship (DPC4), \alphaarchitect v2 automatically designs a three-level prefetcher that outperforms the winning hand-designed solution, further demonstrating automated agentic discovery as a useful tool for computer architects.
Our discovered policy achieves a 3.8\% geometric mean IPC speedup over the baseline overall and a 0.3\% improvement over the prior champion, BertiGO.
On low-bandwidth single-core configurations, our policy yields a 4.6\% performance speedup compared to only 2.6\% for BertiGO.
However, multi-core evolution still remains a significant challenge due to simulation latency impeding evolution speed.
Finally, our profiling of an ArchAgent evolution of over 12,000 candidate designs provides key insights into how automated evolutionary agents explore and synthesize complex microarchitectural logic.

\end{abstract}

\section{Introduction}

\begin{figure}[t]
  \centering
  \includegraphics[width=0.9\linewidth]{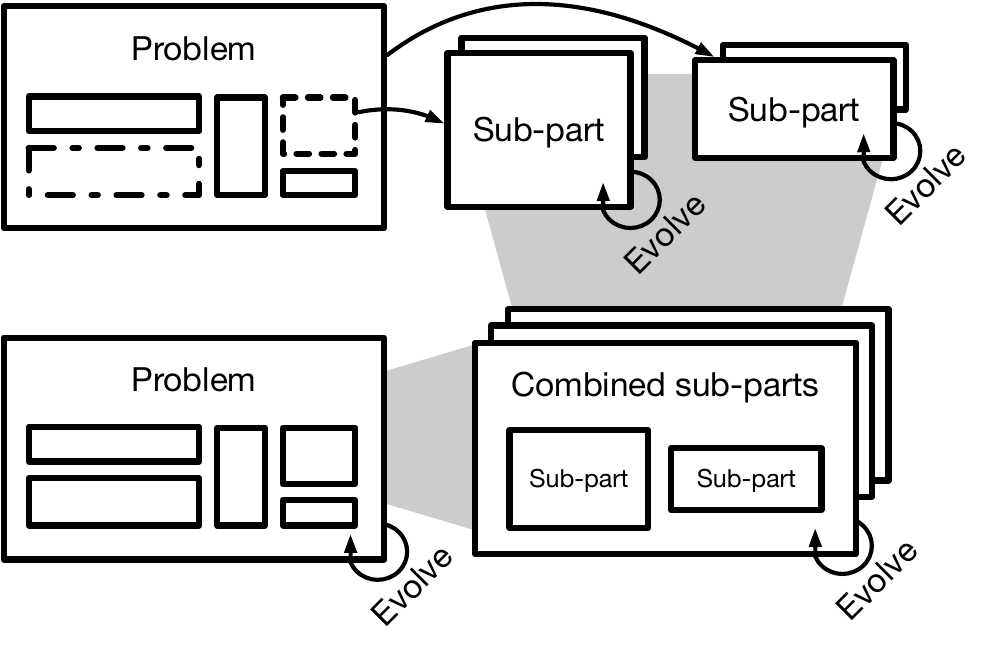}
  \caption{Cascaded divide-and-conquer evolutionary search.}
  \label{fig:divi-conq}
\end{figure}

Recent advancements in agentic artificial intelligence (AI) have driven major breakthroughs in automated algorithm design.
Within the computer systems community, closed-loop machine-learning frameworks are increasingly deployed to automate optimization and ideation.
Specifically, the \alphaarchitect framework~\cite{gupta2026archagent} pioneered autonomous microarchitecture discovery by building a closed-loop evolutionary system for computer microarchitecture.
By coupling a large-language-model-based (LLM-based) code-generation agent (i.e., AlphaEvolve~\cite{novikov2025alphaevolve}) with the ChampSim simulator~\cite{gober2022championship}, ArchAgent iteratively generates hardware policies, evaluates them in trace-based simulation, and uses performance feedback on the evolved policy to guide subsequent mutations.

While \alphaarchitect successfully discovered single-level cache replacement policies, it fails to scale to larger architectural problems, such as multi-level prefetchers, which present higher degrees of freedom and stricter physical hardware constraints.
In particular, applying such systems to multi-level data prefetching increases difficulty along three architectural dimensions. 
First, the design space explodes combinatorially as prefetchers must calculate both target addresses and optimal timing while tracking program and system-level context.
Second, multi-level prefetching introduces an additional dimension of complexity: evolving prefetchers at all three levels simultaneously creates a non-monotonic search space where a performance-improving mutation at one level can completely invalidate the gains at another level.
Finally, prefetchers must adhere to strict, independent physical storage budgets (e.g., 32KB of state at the L1D) that cannot be reliably enforced in existing higher-level simulators such as ChampSim.

To overcome these barriers, we propose \alphaarchitect v2, introducing two primary methodology changes:

\begin{itemize}
    \item {\bf Hardware-Realizability Storage Feedback Loop}: We implement an automated interface that calculates the physical storage of each proposed prefetcher. Over-budget designs are rejected during compilation and are not evaluated, providing negative rewards to the system.
    \item {\bf Cascaded Evolutionary Search}: Rather than evolving all prefetchers simultaneously, we partition the search space. The framework sequentially isolates and optimizes prefetchers at each cache level (L1D, L2, and LLC), using the best results of prior levels to prune the search before conducting global optimizations.
\end{itemize}

We evaluate \alphaarchitect v2 with the setup of the Fourth Data Prefetching Championship (DPC4)~\cite{Home}, starting search from the competition baseline (i.e., Berti~\cite{navarro2022berti} at L1D, Pythia~\cite{bera2021pythia} at L2, no prefetcher at LLC) and optimizing for the official competition score.
To prevent overfitting, the search is performed on the official training workloads, while final prefetching solutions are evaluated on a separate, longer, held-out validation set provided by the competition.

\alphaarchitect v2 automatically designs an effective competition-winning set of L1D, L2, and LLC prefetchers. 
The discovered multi-level prefetching policy achieves a 3.8\% geometric mean IPC speedup over the competition baseline across all workloads and configurations, representing a 0.3\% improvement over the championship-winning, hand-designed policy, BertiGO~\cite{singh2026pushing}.
Notably, in low-bandwidth single-core configurations, where cache pollution and bus contention are most severe, our policy significantly outperforms BertiGO's 2.6\% speedup by an additional 2\%.
Despite large single-core wins, our discovered policies exhibit limited gains in multi-core settings due to 
multi-core feedback being introduced only during the final stage of evolution. 
Multi-core evolutionary search remains challenging for automated systems due to high simulation latencies and an exponential increase in search space complexity.

To understand how evolutionary agents navigate microarchitectural design spaces, we analyze and profile over 12,000 prefetchers generated by \alphaarchitect v2.
We find that major breakthroughs stem from the majority of architectural shifts that require subsequent minor refinements to unlock speedups.
The winning design succeeds by dynamically scaling prefetch lookahead, throttling bandwidth based on multiple core/memory signals to prevent congestion, and intelligently arbitrating between intra- and inter-page memory streams.

In summary, this paper makes the following contributions:
\begin{itemize}
\item We introduce a \textbf{``cascaded" divide-and-conquer evolutionary approach} that sequentially evolves subsets of different cache levels (e.g., L1D, then L2, then LLC).

\item We develop an \textbf{automated feedback loop for a prefetchers footprint}, enforcing strict physical constraints through appropriate feedback.

\item We show that \textbf{\alphaarchitect v2 discovers prefetchers that achieve a 3.8\% performance uplift over the baseline and a 0.3\% improvement over the championship hand-designed policy, BertiGO}, with a 4.6\% improvement over baseline on low-bandwidth systems.

\item We present an \textbf{analysis of over 12,000 evaluated ideas}, exploring one lineage of the winning policy's evolutions, describing how such systems explore microarchitecture.
\end{itemize}

\section{Background}

\begin{figure}[t]
  \centering
  \includegraphics[width=\linewidth]{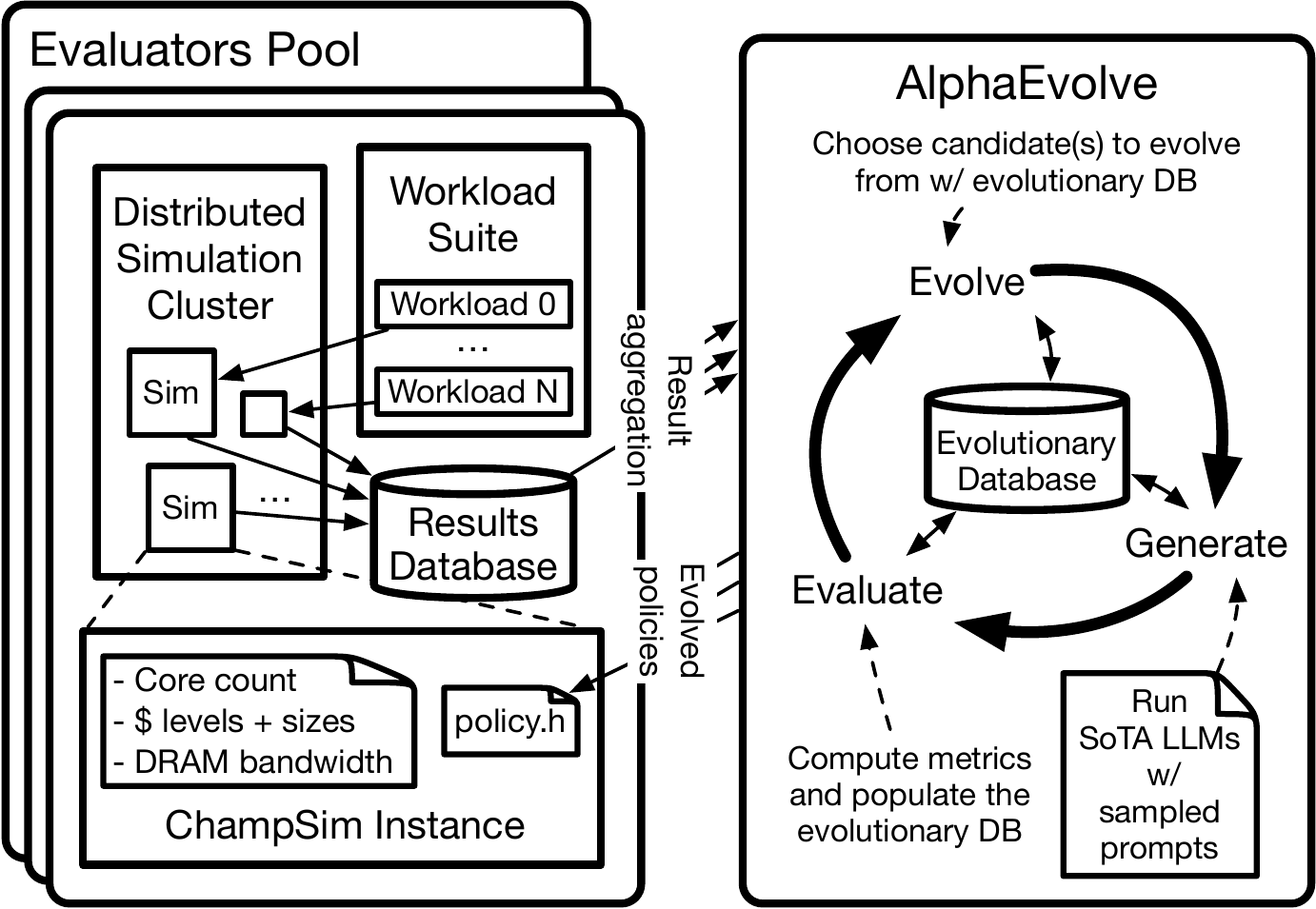}
  \caption{Recap overview of the \alphaarchitect system.}
  \label{fig:aiarch}
\end{figure}

In this section, we first review the general architecture of agentic LLM-based discovery systems.
We then discuss the original ArchAgent~\cite{gupta2026archagent} design, as shown in Figure~\ref{fig:aiarch}, and then outline the DPC4 data prefetching competition rules and setup optimized by \alphaarchitect v2.

\subsection{Primer on Evolutionary Agentic LLM-Based Discovery}

As LLMs have become more capable, a new class of agentic systems (i.e., harnesses) for algorithmic and scientific discovery have grown in popularity.
These discovery systems --- ranging from evolutionary tools like Google DeepMind's \anonevolve~\cite{novikov2025alphaevolve}, SkyDiscover~\cite{liu2026skydiscover}, OpenEvolve~\cite{openevolve}, and ShinkaEvolve~\cite{lange2025shinkaevolve}, to more fully-agentic systems like ScientistOne~\cite{meng2026scientistone} and Glia/Engram~\cite{hamadanian2026gliahumaninspiredaiautomated,karimi2026improvingcoherencepersistenceagentic} --- leverage LLMs to programmatically propose, refine, and compile code solutions optimized for target performance objectives.
Unlike existing coding harnesses or more reasoning-acting coding agents (e.g., Google Antigravity~\cite{GoogleAntigravity}, Anthropic Claude Code~\cite{Claude}, OpenAI Codex~\cite{Codex}, OpenCode~\cite{opencode2026}), these discovery systems typically add more determinism and utilize hardened pre-specified evaluations to ensure a valid optimization over long runtimes.
While initially developed for scientific algorithm discovery~\cite{novikov2025alphaevolve} and software systems optimization~\cite{cheng2025barbarians}, such frameworks have recently been proposed for automated hardware design and architectural co-design~\cite{gupta2026archagent, Venues, blasberg2026agentic, sankaralingam2026computerarchitecturesalphazeromoment}.

Focusing on evolutionary frameworks (i.e., ~\cite{novikov2025alphaevolve,liu2026skydiscover,openevolve,lange2025shinkaevolve}), the core of these frameworks is an orchestration loop that continuously ``evolves'' code.
The controller first samples from a structured program database --- frequently managed via quality-diversity algorithms like MAP-Elites~\cite{mouret2015illuminatingsearchspacesmapping} or island-based populations~\cite{journals/nature/RomeraParedesBNBKDREWFKF24,tanese1989distributed} --- using prior implementations and their associated metrics to seed subsequent iterations.
The generation stage employs an LLM, either as a single-turn generator or a multi-turn coding agent, to synthesize new candidate implementations.
These candidates are then evaluated using a user-defined method and the resulting performance metrics are returned to the controller to update the program database.
This iterative loop of sampling, generating, and evaluating continues until the optimization target or search timeout is reached.
To guide the generation step, users provide a base prompt detailing the target problem, the execution interface, and any domain-specific hints to guide the search.

\subsection{\alphaarchitect v1 Overview}

The ArchAgent framework~\cite{gupta2026archagent} (ArchAgent v1) was the first-of-its-kind automated computer architecture discovery system built on AlphaEvolve~\cite{novikov2025alphaevolve}.
Initially targeting cache replacement, ArchAgent focused on discovering new C++ model implementations of cache replacement policies compatible with the ChampSim microarchitectural simulator.

To achieve this, the framework constructs LLM prompts that establish an expert computer architect persona, outline the ChampSim cache replacement API and class structure, and define strict hardware storage constraints (e.g., a 48KB budget for cache replacement metadata).
Each generated policy is subsequently evaluated in parallel across a comprehensive set of workload traces.
The simulator measures performance in instructions per cycle (IPC) across these target workloads.

Using this closed-loop evolutionary process, ArchAgent automatically discovers high-performing policies within a month of autonomous search, achieving a 0.9\% IPC speedup over the state-of-the-art on heavily optimized single-core SPEC 2006 workloads, and over a 5.3\% IPC speedup on public multi-core Google Workload Traces~\cite{Google_Workload_Traces_Version_2}.

\subsection{Fourth Data Prefetching Championship and BertiGO}
\label{sec:dpc4}

Established in 2009, the Data Prefetching Championships (DPC)~\cite{Home, Hparch, alameldeenfirst, Pico} aim to advance the state-of-the-art in data prefetching by providing the community with standardized infrastructure and traces to evaluate prefetching ideas.
The fourth iteration of the championship, DPC4~\cite{Home}, was held in 2026, and it built upon prior iterations of the competition by expanding the scope of multi-level prefetching (at the L1D, L2, and LLC), updated workloads (spanning AI/ML inference, datacenter-class workloads, complex graph-based apps), and more core configurations (single- and multi-core configurations).

Prefetchers are evaluated on ChampSim, a widely-used academic C++-based microarchitectural simulator.
Participants implement prefetchers utilizing common APIs to access microarchitectural state and system-level feedback where designs must adhere to strict, independent physical storage limits per cache level: 32KB for L1D, 128KB for L2, and 256KB for LLC.
To ensure solutions generalize to different operating conditions, submissions are tested across three distinct system configurations: two single-core configurations (one with ample and one with restricted DRAM bandwidth) and a multi-core configuration with ample DRAM bandwidth.
Training workloads are provided to participants, but final submissions are ranked based on their performance over a separate, held-out evaluation workload suite.

All championship submissions are measured against a state-of-the-art baseline consisting of the Berti prefetcher at the L1D level, a Pythia prefetcher at the L2 level, and no prefetcher at the LLC.
The state-of-the-art (SoTA) DPC4 submission, BertiGO~\cite{singh2026pushing}, builds on this baseline by introducing path signature tracking to the L1D Berti engine, integrating set-dueling mechanisms into the L2 Pythia prefetcher, and adding an adaptive next-line prefetcher at the LLC.
The majority of BertiGO's gains are driven by L1D optimization, with only minor performance gains contributed by L2 and LLC updates.

\section{\alphaarchitect v2 For Data Prefetching}
\label{sec:archagent-dpc}

To discover multi-level prefetchers, we extend \alphaarchitect in two ways: by adding (1) a hardware-realizability feedback loop, and (2) a cascaded divide-and-conquer methodology to break down the problem into smaller tractable problems.

\subsection{Hardware-Realizability Storage Feedback Loop}

\begin{figure}[t]
\centering
\begin{lstlisting}[
  language={}, % Disables syntax highlighting
  basicstyle=\ttfamily\footnotesize,
  backgroundcolor=\color{gray!5}, % Light gray background
  frame=single,                   % Single frame around the text
  rulecolor=\color{black},        % Black border
  breaklines=true,                % Auto-wrap lines if they exceed width
  breakatwhitespace=true,
  columns=fullflexible,
  keepspaces=true
]
The simulated storage budgets for the prefetcher
to implement is:
  - L1D `my_l1d_prefetcher` = 32KB
  - L2 `my_l2_prefetcher` = 128KB
  - LLC `my_llc_prefetcher` = 256KB
  
Honestly stay within the budget and truthfully
report the number of utilized bytes using the
`int prefetcher_size()` function per prefetcher.
When you add a new component, you should account
for its size in the `prefetchersize()` function.
You should only remove an entry from the
`prefetcher_size()` function when you remove the
component.
If a C++ type is used to represent hardware,
ensure proper masking is added to enforce that the
C++ type is used correctly.
For example, if a `uint64_t` is used to represent
3 bits, any write to that C++ symbol using
`uint64_t` must mask out the upper 61 bits
(64 - 3) before the write.
In this example, the 3 bit size can be used for
prefetcher size budget calculations instead of the
full 64 bit size of the C++ type.

FALSE REPORTING OR EXCEEDING THE STORAGE BUDGET
WILL BE REJECTED.
\end{lstlisting}
\caption{Example storage budget restriction prompt.}
\label{fig:storage_budget}
\end{figure}

While the ChampSim simulator provides mechanisms to measure metrics such as IPC, it does not provide a mechanism to measure the hardware cost of proposed prefetchers (i.e., area or power costs). 
Typically for competitions such as DPC4 which require strict storage budgets (32KB for L1D, 128KB for L2, and 256KB for LLC prefetchers), this area cost is estimated after the algorithm is complete by hardware architects manually estimating the size of different components and obtaining an overall size.
However, a major challenge with automated code generation is that LLM optimizers often optimize solely for performance metrics, proposing complex algorithms that violate physical hardware constraints.

To enforce these competition storage budgets during evolution, we integrated hardware-realizability area feedback in \alphaarchitect.
First, like \alphaarchitect v1, we prompted the LLM with the precise storage limits so that ideas would probabilistically adhere to storage limits.
Next, we implemented a custom simulator interface function per prefetcher, \texttt{prefetcher\_size()}, which calculates the policy's storage footprint based on its state and returns an integer size for bytes of storage.
We used initial literature to fill out this function per baseline prefetcher (e.g., Berti and Pythia), using C++ parameters and code to derive the overall size at runtime.
During compilation and simulation, the evaluation checks the \texttt{prefetcher\_size()} function per prefetcher and checks against the competition size requirements.
If a candidate policy exceeds the allowed budget, it is rejected immediately.
Finally, we updated the prompt as shown in Figure~\ref{fig:storage_budget} to guide the LLM to update both the prefetcher logic and the prefetcher size function, ensuring size calculations remain synchronized with implementation changes.
Through offline LLM judges, the code generation step of \alphaarchitect was able to faithfully adhere to updating both logic and size functions properly, while a manual check was done on the final output to ensure sizing was correct.
Combined these additions allowed the LLM to better adhere to the storage budget per prefetcher throughout the evolution process bounding viable ideas by one aspect of realizability.

\subsection{Cascaded Evolutionary Search}

\begin{figure}[t]
  \centering
  \includegraphics[width=\linewidth]{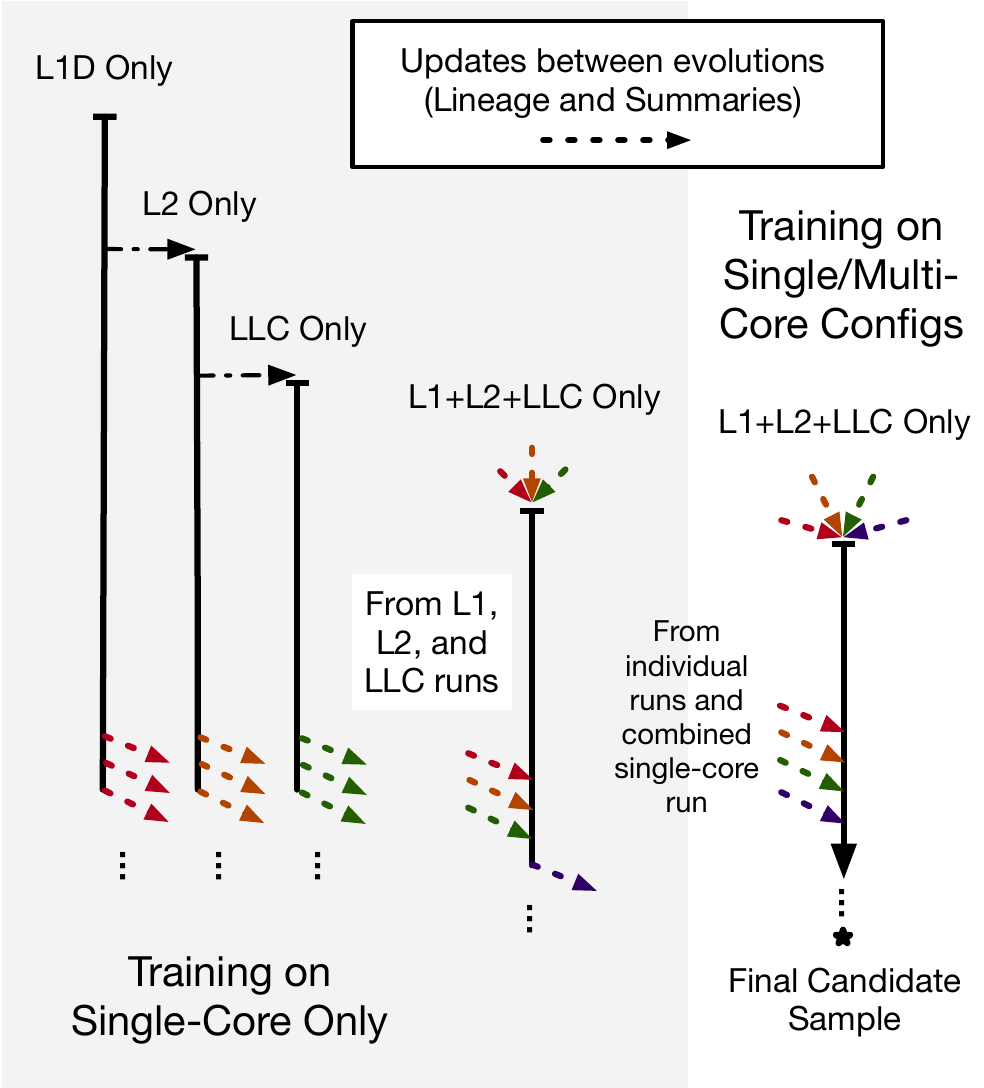}
  \caption{Evolutionary search ``cascade'' methodology.}
  \label{fig:testing}
\end{figure}

The combination of optimizing three prefetchers in-tandem and slow ChampSim simulations (up-to or exceeding 12 hours to evaluate each change) introduces an extremely large and slow search space to navigate, leading to \alphaarchitect v1 struggling to make meaningful progress.
To address this, we split the design space into a ``cascade''-style divide-and-conquer evolutionary process where each prefetcher was evolved in isolation, then globally-optimized across all prefetchers, in multiple stages.
Furthermore, we reduced the design space by limiting the search to optimize only single-core configurations when evolving individual prefetchers.
Only when all prefetchers were optimized, did we introduce multi-core configurations into the search space.
This methodology is shown in Figure~\ref{fig:testing}.

This strategy can be broken down into two general phases: first optimizing each prefetcher level in sequence, then globally-optimizing all prefetchers.

\subsubsection{Individual Prefetcher Optimizations}

To begin with, we evolved each prefetcher in isolation, re-using the best result from a lower level of the hierarchy as an initialization for the next level. 
In addition, we only evolved with single-core ChampSim workload configurations, equally weighing the low and high bandwidth configurations.
The setup was as follows:

\begin{enumerate}
  \item \textit{L1D optimization}: We first evolved only the L1D prefetcher (starting from Berti), keeping the baseline L2 (i.e., Pythia) and LLC (i.e., nothing) prefetchers static.
  \item \textit{L2 optimization}: Next, once the L1D prefetcher performance began to plateau, we selected the best-performing L1D candidate, froze its implementation, and then evolved only the L2 prefetcher (starting from Pythia).
  \item \textit{LLC optimization}: After the L2 optimization converged, we then froze the best L1D and L2 prefetcher candidates, and then evolved only the LLC prefetcher.
\end{enumerate}

While this selectively optimized for L1D, then L2, then LLC, this matched prior intuition that workloads were more amenable to L1D prefetcher optimization (as seen in DPC4 submissions).
This L1D dominant optimization matched the prior winner, BertiGO's, improvements, which were L1D dominant. 

\subsubsection{Globally-Optimizing All Prefetchers}

While the individual prefetcher optimizations were running, two new evolutionary runs were spawned to do global optimizations of the algorithms presented.
The goal of these runs was to further optimize all prefetchers together instead of individually to allow for synergistic improvements between prefetchers.
This was done by a prior runs database, which stored previous successful bests from any evolutionary run, in two parts:

\begin{enumerate}
  \item \textit{Single-core cross-prefetcher optimization}: First, all prefetchers were optimized at once with the seed best prefetchers from each individual optimization run only on single-core ChampSim workloads.
    In addition, at random intervals, the prompt to generate new prefetcher candidates was injected with the top performing candidates and their lineages/summaries from the other individual optimization runs (e.g., only the L1D) while they were running.
    This allowed upstream individual prefetcher runs to continue running and optimizing while influencing the combined evolution.
    Additionally, in this run, we double weighted the low bandwidth single core configuration (e.g., \texttt{geomean(lowBW, lowBW, highBW)}) to encourage evolution to optimize low bandwidth configurations.
    This was done to give a proxy for multi-core workloads, avoiding slow multi-core simulation speeds, since the multi-core configuration splits its bandwidth across multiple cores similar to the low bandwidth single-core configuration.
  \item \textit{Multi-core cross-prefetcher optimization}: Then to produce a final set of prefetcher candidates, we then took the best prefetchers from the cross-prefetcher optimization, and further evolved them concurrently with both the single-core and multi-core configurations.
    Similar to the other cross-prefetcher optimization, at random intervals, the prompt was injected with top performing candidates from all other runs (both the individual prefetcher optimizations and the single-core cross-prefetcher optimizations).
\end{enumerate}

This strategy reduces the complexity of LLM reasoning per step and prioritizes search effort on the near-to-core cache levels, where there is highest latency hiding potential.

\subsection{Training/Evolution Versus Validation Methodology}

\begin{table}[t]
\centering
\caption{Training/evolution and simulation methodology.}
\label{tab:simulation_parameters}
\small
\begin{tabularx}{\columnwidth}{l l Y Y}
\toprule
\multirow{2}{*}{\textbf{Configuration}} & \multirow{2}{*}{\textbf{Phase}} & \multicolumn{2}{c}{\textbf{Instructions (Millions)}} \\
\cmidrule(lr){3-4}
 & & \textbf{Total} & \textbf{Warmup} \\
\midrule
\multirow{2}{*}{Single-Core} & Training/Evolution & 60 & 15 \\
\cmidrule{2-4}
 & Full Validation & 200 & 50 \\
\midrule
\multirow{2}{*}{Multi-Core} & Training/Evolution & 12.5 & 5 \\
\cmidrule{2-4}
 & Full Validation & 50 & 20 \\
\bottomrule
\end{tabularx}
\end{table}

We used the existing competition benchmarks to evaluate each candidate policy.
Specifically, we evaluated candidates on a reduced instruction trace length during the evolutionary search phase (training/evolution) and validated the final selections on long instruction traces to ensure optimality and generality.

Table~\ref{tab:simulation_parameters} describes the training and evaluation methodology used.
During evolution, simulations ran for 60 million instructions after 15 million instructions of warmup for single-core, and 12.5 million instructions after 5 million warmup for multi-core.
In contrast, full validation ran for 200 million instructions (50 million warmup) and 50 million instructions (20 million warmup), respectively.
Single-core benchmarks were evaluated individually according to the championship guidelines.
For multi-core configurations, during training, we generated 100 random combinations of mixed workloads (e.g., SPEC17 and Graph workload mixes) to evaluate the policies under diverse and heterogeneous resource contention.
The final evaluation used the championship evaluation set of single- and multi-core traces, which included both unseen workloads for single-core configurations and novel trace mixes for multi-core configurations to verify generalizability.

This mixture of trace lengths and workloads for both training and validation ensured robust generated policies.
We expect that expanding the training set to cover additional workloads would yield even greater robustness.

\section{The Evolved Multi-Level Prefetching Algorithm}

We describe and evaluate the winning generated multi-level prefetcher policy, using the Section~\ref{sec:archagent-dpc} methodology, that beats DPC4 SoTA policies within two months of evolution time.

\subsection{Prefetcher Descriptions}

In this subsection, we describe in detail the L1D, L2 and LLC prefetcher components. The generated prefetcher ensemble follows two key design principles. First, it adds \textbf{new engines to track patterns that the baseline did not capture, such as, global strides, irregular deltas, and temporal prefetches}. Secondly, it introduces \textbf{new arbitration mechanisms to choose prefetch engines and modulate aggressiveness}.

Each prefetcher adheres to the competition storage budget for each cache; 31.1 KB for L1D (out of 32 KB), 110.0 KB for L2 (out of 128 KB), and 230.3 KB for LLC (out of 256 KB).
Table~\ref{tab:storage} further breaks down the storage used, per component, for each prefetcher.

\subsubsection{L1D Prefetcher}

\begin{figure}[t]
  \centering
  \includegraphics[width=0.9\linewidth]{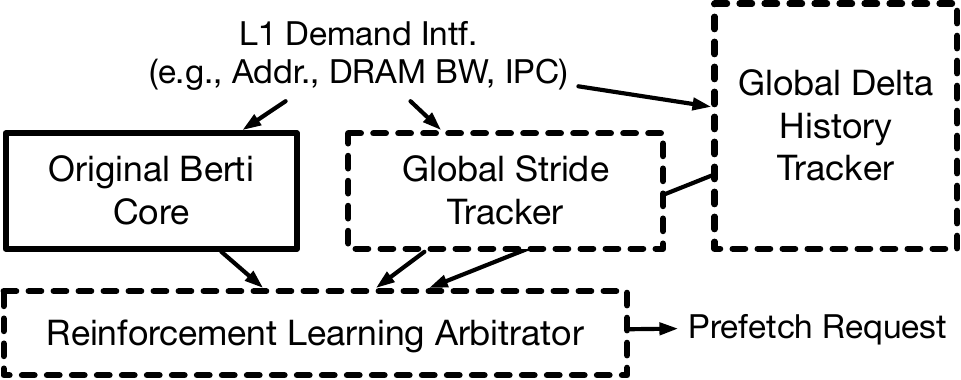}
  \caption{High-level overview of the L1 modifications. Dotted lines indicate new/modified components.}
  \label{fig:l1}
\end{figure}

\begin{table}[t]
\centering
\normalsize
\caption{Storage breakdown for the evolved prefetchers.}
\spaceskip=0.15em 
\label{tab:storage}
\renewcommand{\arraystretch}{1.2}
\setlength{\tabcolsep}{4pt}
\begin{tabular}{llrr}
\hline
& \textbf{Structure} & \textbf{Breakdown} & \textbf{KB} \\ \hline
\multirow{11}{*}{\textbf{L1D}} 
 & Current Pgs. Tbl. & 63e $\times$ 277b & 2.1 \\
 & Prev. Requests Tbl. & 1024e $\times$ 28b & 3.5 \\
 & Prev. Prefetches Tbl. & 512e $\times$ 57b & 3.6 \\
 & Record Pgs. Tbl. & 1024e $\times$ 135b & 16.9 \\
 & IP Tbl. & 1024e $\times$ 10b & 1.3 \\
 & Global Stride Tracker & 128e $\times$ 110b & 1.7 \\
 & Global Delta Hist. Tbl. & 512e $\times$ 20b & 1.3 \\
 & RL. Arbiter & 512e $\times$ 12b & 0.8 \\
 & Misc. Registers & -- & 0.04 \\ \cline{2-4}
 & \textbf{Total L1D} & \textit{\textbf{Limit: 32KB}} & \textbf{31.1} \\ \hline
\multirow{4}{*}{\textbf{L2}} 
 & Q-Value Store & 5fts. $\times$ 18acts. $\times$ 16b & 99.0 \\
 & Signature Tbl. & 128e $\times$ 300b & 4.7 \\
 & Prefetch Tracker & 128e $\times$ 403b & 6.3 \\ \cline{2-4}
 & \textbf{Total L2} & \textit{\textbf{Limit: 128KB}} & \textbf{110.0} \\ \hline
\multirow{4}{*}{\textbf{LLC}} 
 & Stride Tbl. & 2048e $\times$ 67b & 16.8 \\
 & Tagged Address Tbl. & 12288e $\times$ 41b & 61.5 \\
 & Global Event History & 32768e $\times$ 38b & 152.0 \\ \cline{2-4}
 & \textbf{Total LLC} & \textit{\textbf{Limit: 256KB}} & \textbf{230.3} \\ \hline
\end{tabular}
\end{table}

The evolved L1D prefetcher, as shown in Figure~\ref{fig:l1}, enhances the baseline Berti design by incorporating a Global Stride Tracker (GST) and Global Delta History Table (GDHT) to complement Berti's local-delta tracking system. Additionally, a Reinforcement Learning (RL) Arbitrator is added to select prefetch candidates across the engines.

{\it Global Engines}: While the baseline Berti prefetcher is effective at capturing local, per-IP patterns within page boundaries, it does not capture cross-IP spatial streams that traverse physical pages. To cover this gap, two engines were added: the Global Stride Tracker (GST) covers linear step streams, and the Global Delta History Table (GDHT) covers irregular page-level jumps that don't follow a regular step pattern.

{\it Arbitration}: An RL Arbitrator was introduced to choose between candidates from the three different prefetch engines.
The arbiter uses a global path, from the PC and a sliding delta history context window to evaluate different engines.
Additional metadata is added to link long-latency evictions directly back to the specific engine that issued them, allowing the arbiter to accurately reward or penalize the GST and GDHT engines based on real-time cache utility.
To prevent bus congestion and cache thrashing, real-time memory bandwidth thresholds were also used to dynamically throttle the arbitrator's aggressiveness. 

\subsubsection{L2 Prefetcher}

\begin{figure}[t]
  \centering
  \includegraphics[width=0.9\linewidth]{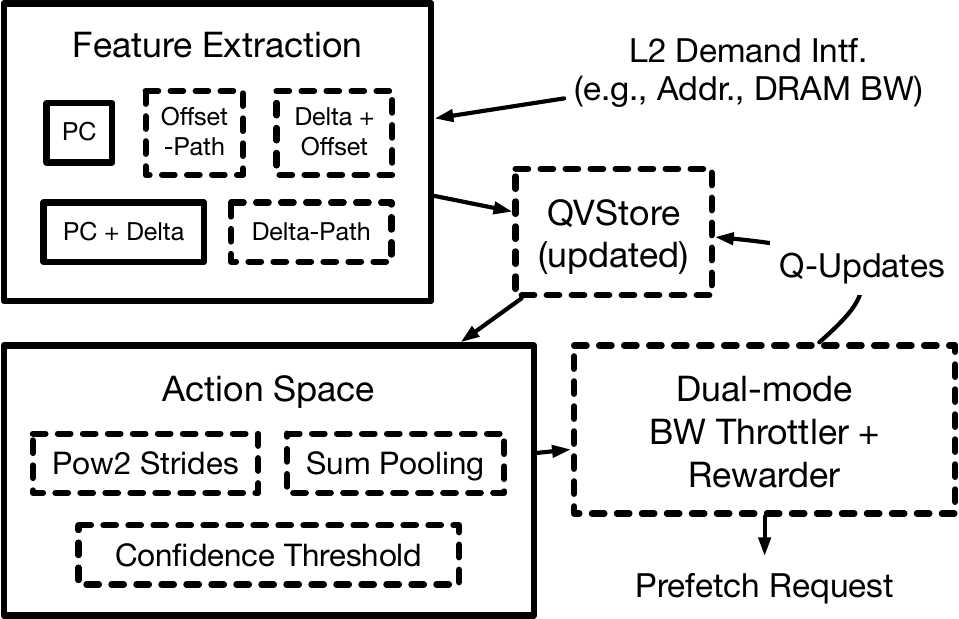}
  \caption{High-level overview of the L2 modifications. Dotted lines indicate new/modified components.}
  \label{fig:l2}
\end{figure}

The evolved L2 prefetcher refines the Pythia baseline without major changes, focusing on complementing the L1D prefetcher. Rather than large-scale restructuring, ArchAgent upgraded the RL feature representations and action-space characteristics as shown in Figure~\ref{fig:l2}.

{\it Context, Action, and Pooling Upgrades}: Pythia's RL framework was upgraded with extra features (from 2 to 5 features) and extending available action step choices to 18 choices (from 15).
The 5 active features expanded to cover local program context (e.g., using only one PC or PC with a delta) as well as PC history (e.g., using the last 4 PCs, deltas, and offsets).
The 18 action choices are created by hardcoding the action space to mainly generate positive and negative power-of-two memory strides helping to cover many variations of streaming, strided, and matrix access patterns.
Finally, ArchAgent switched to sum-pooling with a confidence threshold, to help reduce cache pollution and DRAM traffic during unfamiliar or non-strided access patterns.

{\it Bandwidth Throttling}: Under high memory traffic, the throttler limits prefetch burst degrees and enforces higher confidence to send a prefetch.
Additionally, the reward feedback is augmented in these high traffic states, to force less bus congestion.

\subsubsection{LLC Prefetcher}

\begin{figure}[t]
  \centering
  \includegraphics[width=0.9\linewidth]{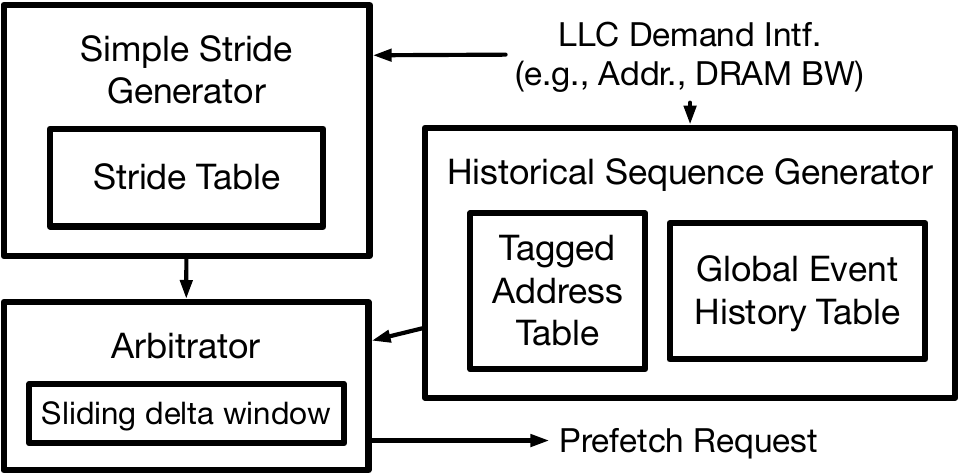}
  \caption{High-level overview of the LLC modifications. Dotted lines indicate new/modified components.}
  \label{fig:llc}
\end{figure}
\begin{figure*}[t]
  \centering
  \includegraphics[width=\linewidth]{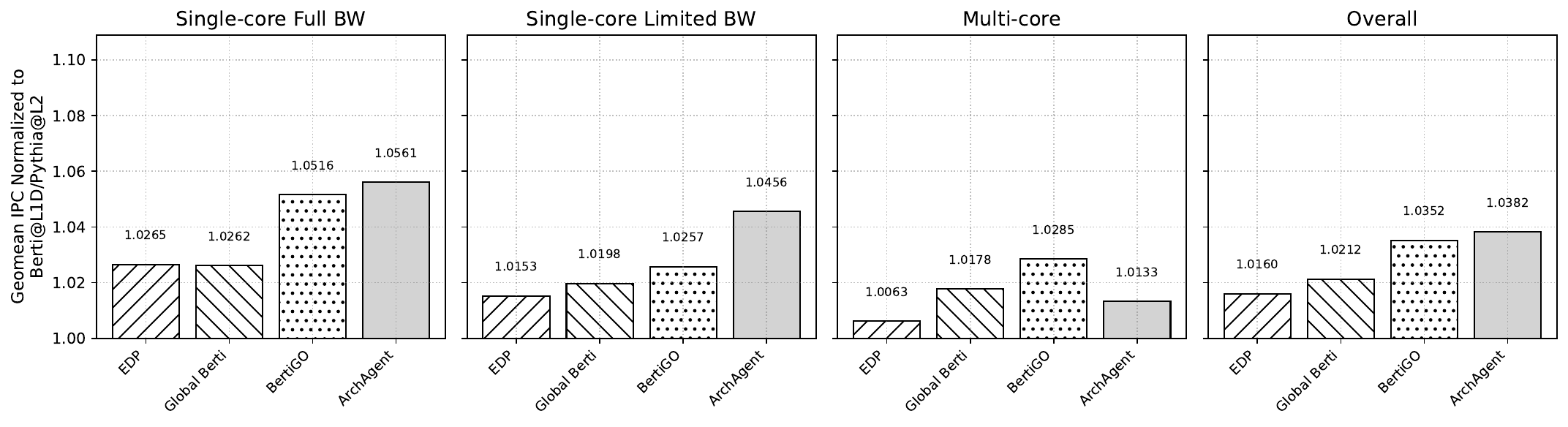}
  \caption{\alphaarchitect's policy compared against the baseline/SoTA prefetchers (i.e., EDP~\cite{navarroentangling}, Global Berti~\cite{posluns2026global}, and BertiGO~\cite{singh2026pushing}).}
  \label{fig:scores-plot}
\end{figure*}
\begin{figure*}[t]
  \centering
  \includegraphics[width=\linewidth]{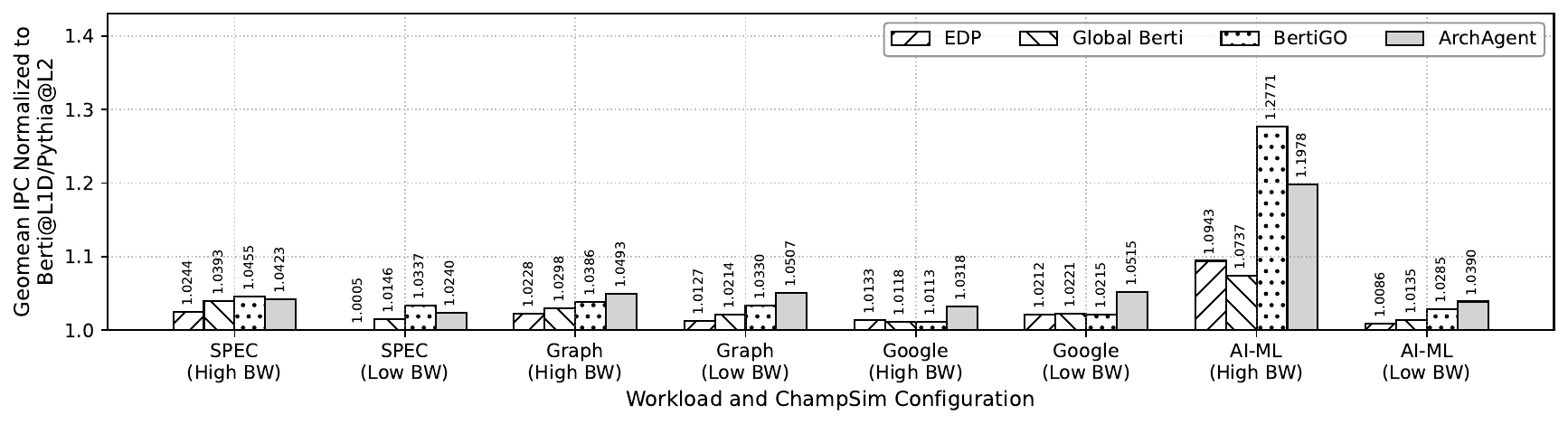}
  \caption{\alphaarchitect's policy measured across distinct workload suites and single-core ChampSim championship configurations.}
  \label{fig:single-core-suites}
\end{figure*}
\begin{figure*}[t]
  \centering
  \includegraphics[width=\linewidth]{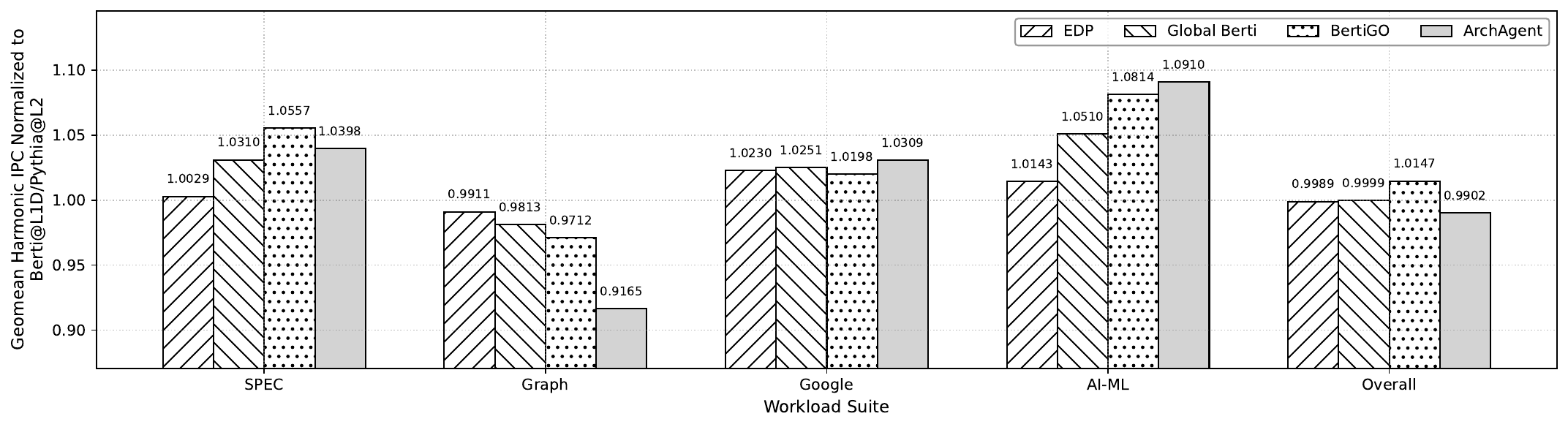}
  
  \caption{\alphaarchitect's policy measured across distinct workload suites using the multi-core ChampSim championship configuration.}
  \label{fig:multi-core-suites}
\end{figure*}

At the LLC level, two prefetch generators were added as shown in Figure~\ref{fig:llc}.
The first prefetch generator records and replays irregular, pointer-chasing miss sequences.
The second prefetch generator detects regular linear and strided access patterns.
Both prefetch generators modulate prefetch degree based on real-time DRAM bandwidth as well.
These are arbitrated by a sliding delta tracking window which monitors recent cache memory misses and calculates the mathematical ratio of unique step distances.
If memory requests are orderly and follow simple repeated steps, it uses the standard regular linear and strided engine.
If requests are highly ``chaotic'', it switches to the historical sequence engine.

\subsection{Evaluation}

Figure~\ref{fig:scores-plot} shows the DPC4 scores for prior multi-level SoTA prefetchers compared to the \alphaarchitect's evolved prefetcher (i.e., prefetching policy).
Overall, compared to the baseline, we see a 3.8\% performance uplift in geomean IPC speedup versus the baseline, winning the competition.
When compared directly to the prior SoTA DPC4 championship winner, BertiGO, we see a 0.3\% uplift.
Looking at the different core count and bandwidth configurations, we see that the performance improvement overall comes from better single-core performance at the cost of multi-core performance.
This is due to the cascade prioritizing single-core performance in the workflow then later optimizing multi-core performance.
Gains over BertiGO come from this optimization for single-core workloads, especially in low bandwidth settings (beating SoTA by over 2\%) at the cost of multi-core performance (where the AI-generated policy loses by about 1.5\%).
In this diagram, multi-core performance evaluations are run using 4-core mixes per suite (i.e., 4 traces all from one suite, e.g., SPEC17) and across suites (i.e., 4 traces with 1 per suite, e.g., 1 from SPEC17, Graph, Google, and AI-ML).

Figure~\ref{fig:single-core-suites} shows the performance of the single-core configurations divided into workload suites.
Here we see that the evolved multi-level prefetcher does better in Graph and Google workloads across low- and high-bandwidth configurations than prior SoTA winners while having mixed gains over BertiGO in other workloads such as AI-ML and SPEC.
Generally, winning margins are higher for low-bandwidth configurations for the evolved policy.

Figure~\ref{fig:multi-core-suites} shows the performance of the multi-core configurations divided into different workloads.
In this diagram, the ``Overall'' grouping represents only using 4-core mixes across suites while suite-level evaluations are run with 4-core mixes within the same suite.
Here the improvements come from Google and AI-ML workloads over prior SoTA, despite the final prefetcher losing in the overall mix.

\section{Profiling an ArchAgent v2 Evolution}

To understand the evolution trajectory of the prefetcher, this section extensively profiles the first---and longest running evolution---for the L1D prefetcher on single-core workloads.
Importantly, this evolution is just the first step of our cascaded approach, so the prefetcher produced at the end of this evolution does not correspond to the L1D prefetcher evolved at the end of all the cascaded steps.
We answer the following questions:

\subsection{How many prefetchers were evaluated?}

\begin{figure}[t]
  \centering
  \includegraphics[width=\linewidth]{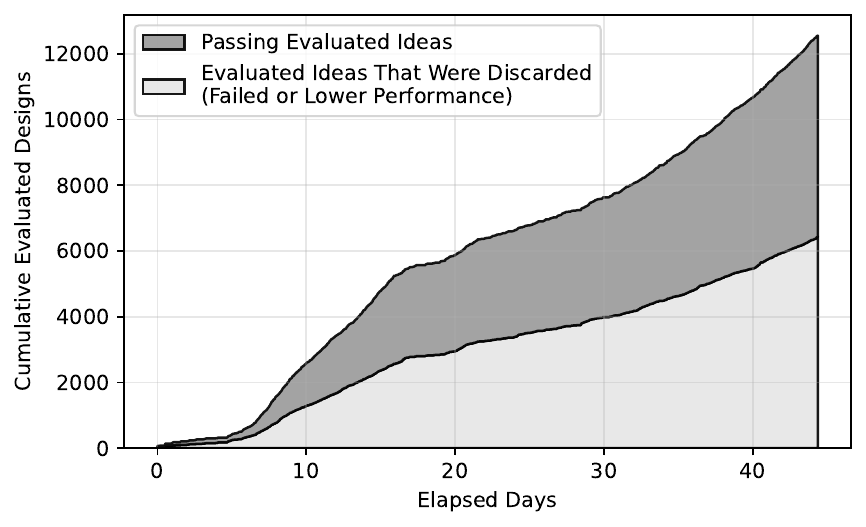}
  \caption{Success and failure of evaluated ideas.}
  \label{fig:success-failures}
\end{figure}

Figure~\ref{fig:success-failures} shows that over 12,000 candidate prefetchers were evaluated during the first evolution's lifetime. These are split into two main categories: prefetchers that were valid and contributed to an improvement, and prefetchers that didn't contribute to an improvement (including designs that compiled incorrectly, had low performance, etc). For this particular evolution, ideas were equally split amongst these two categories. Given the scale of the evolution, we conclude that \textbf{\alphaarchitect could, in parallel, iterate through many more ideas than a human could}. However, the large failure rate also points to inefficiencies in the evolution loop.

\subsection{How varied were the generated prefetchers?}

\begin{figure}[t]
  \centering
  \includegraphics[width=\linewidth]{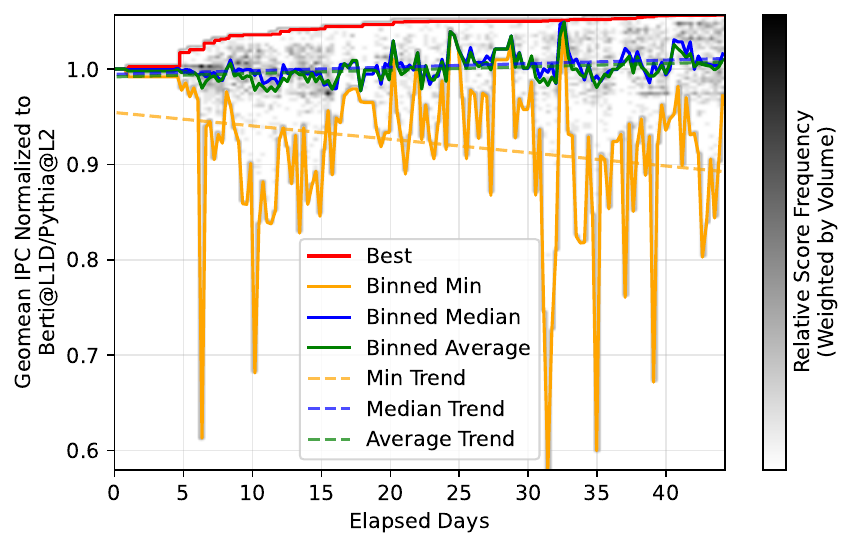}
  \caption{Score and heatmap over time.}
  \label{fig:metric-heatmap}
\end{figure}

Figure~\ref{fig:metric-heatmap} shows the performance variance of designs over time.
The Y-axis shows geomean IPC of the proposed prefetcher compared to the competition baseline (Berti at the L1D and Pythia at the L2), while the X-axis shows time.
The red line is the best result from the evolution at any point in time, and the heatmap (light grey dots) underneath shows evaluations done throughout the evolution lifetime. The point intensity tells how many evaluations led to that same metric at that time.

The heatmap shows that \textbf{designs show high variance in performance}. Despite the overall best increasing over time, \alphaarchitect v2 still tries exhaustively many ideas that range from good to bad. The median/mean trend lines show that the average proposals are still spread around the baseline.
This indicates that while \textbf{the probability of generating a good idea increases over time}, the trajectory of the average proposal does not improve, which again points to inefficiencies in the evolutionary loop. 

\subsection{How did the agent arrive at these final ideas?}

Next, we track the agent's trajectory to understand how it arrived at the final best L1D prefetcher algorithm result for this evolution.  We will see that many ideas and their origins in this evolution contributed to the final prefetcher design.

Since it is impractical to describe and track thousands of prefetcher designs, we define a taxonomy of design modifications that the agent makes broken into ``What'' the LLM did to obtain the performance gain and ``Why'' it deemed to be viable/valuable.
These types of taxonomies are similar to other works that help to diagnose why evolutionary algorithms fail~\cite{pelleriti2026evolutionarycodingagentsevolve, cemri2026multi}.
We reuse the ``What'' taxonomy provided by \cite{pelleriti2026evolutionarycodingagentsevolve} while the ``Why'' taxonomy is created by an offline analysis of all ideas, as shown in Tables~\ref{tab:what}~and~\ref{tab:why}.
For the following analysis, an LLM judge analyzed the specific change made by the agent and attributed a percentage to each taxonomy category (e.g., 50\% attributed to an ``architectural change'' v.s. 50\% to ``hyperparameter tuning'').

\begin{table*}[t]
\normalsize
\centering
\caption{``What'' taxonomy.}
\label{tab:what}
\begin{tabular}{ll}
\toprule
\textbf{Category} & \textbf{Description} \\
\midrule
Bug Fix & Correcting errors or restoring intended behavior without adding features. \\
External Dependency & Changes driven by external libraries, environments, or API surfaces. \\
Architectural Change & Fundamental changes to layout, topology, engine framework, or core logic pipelines. \\
Composition & Combining multiple existing modules or prefetchers to work together. \\
Local Refinement & Tuning or fine-tuning existing placement, mapping, or logic constraints without new structures. \\
Pruning & Removing unused tables, feedback logic, dead code, or redundant components. \\
Refactor & Code restructuring without changing execution performance or underlying logic. \\
Efficiency & Optimizations targeted at cycle latency, execution throughput, or overhead reduction. \\
Hyperparameter Tuning & Adjustments to constant threshold values, decay rates, or static parameters. \\
Other & Any classification that does not strictly fit the structural modifications defined above. \\
\bottomrule
\end{tabular}
\end{table*}

\begin{table*}[t]
\centering
\normalsize
\caption{``Why'' taxonomy.}
\label{tab:why}
\begin{tabularx}{\textwidth}{>{\hsize=0.33\hsize\raggedright\arraybackslash}X L}
\toprule
\textbf{Category} & \textbf{Description} \\
\midrule
Bandwidth Management & Adjusting prefetch aggressiveness based on memory traffic to avoid congestion. \\
Execution History Disambiguation & Using execution history/paths to separate overlapping access patterns.\\
Pattern Detection Expansion & Upgrading logic to capture longer, wider, or inter-page access patterns. \\
Cache Pollution Mitigation & Adding confidence filtering to stop inaccurate prefetches from displacing useful data. \\
Multi-Level Coordination & Making all prefetchers work together rather than conflicting/duplicating effort. \\
Hardware Efficiency & Optimizing logic and storage to fit within physical area and timing constraints. \\
Algorithmic Tuning & Adjusting parameters, counters, or thresholds without changing the core design. \\
Other & Any classification that does not strictly fit the motivations defined above. \\
\bottomrule
\end{tabularx}
\end{table*}

\subsubsection{Breakdown Of All Ideas}

Across all ideas that contributed to the final idea, we see that 50\% of changes were ``architectural changes'' and there is a long tail of other changes that are in other categories.
This shows that \alphaarchitect generates various kinds of changes that are intermingled into more performant changes.
Analyzing ``why'' it produced the changes led to a more even mix.
First, we see that ``pattern detection'' is dominant at about 35\% of all ideas.
Of the remaining ideas, most ideas were spawned from the LLM thinking about ``hardware efficiency,'' ``execution history disambiguation,'' ``bandwidth management,'' and ``cache pollution,'' each about 14\%.
This shows that the LLM, given the feedback from the simulator, was able to infer enough about the workloads, execution history, and more to make reasonable optimizations.
We expect that with even more context provided (e.g., more metrics or workload analysis), we can further refine what and why a change was made.

\subsubsection{Main Lineage Breakdown}

\begin{figure}[t]
  \centering
  \includegraphics[width=\linewidth]{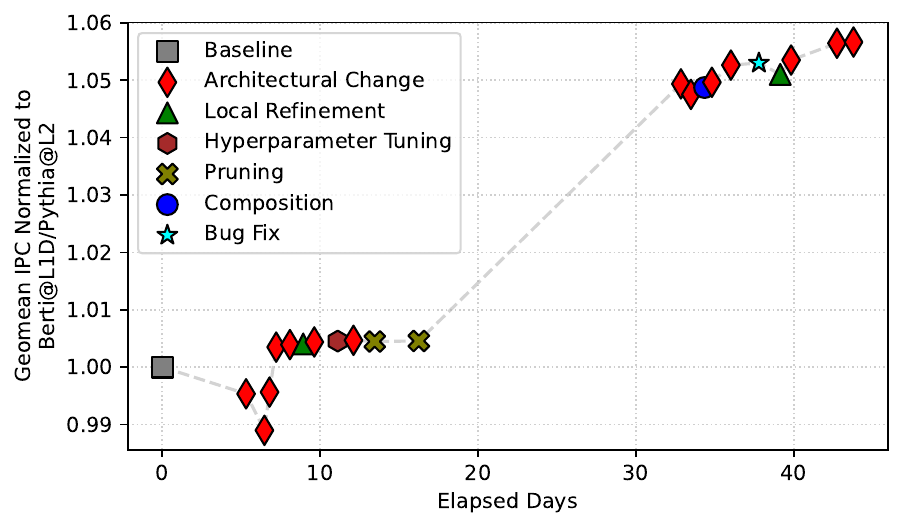}
  \caption{Annotated main lineage.}
  \label{fig:main-lineage}
\end{figure}

Next, we look at the ideas along the main lineage, which is defined as the evolutionary path of the best program back to the initial program baseline.
Figure~\ref{fig:main-lineage} shows the breakdown of the main lineage with respect to the ``What'' taxonomy.
If a point had multiple categories associated with it, we only display the dominant one.
Note, the primary lineage doesn’t have monotonically increasing performance.
This shows that an idea might not produce any performance benefit until it is refined or composed with other changes.

For the main lineage, we see that 65\% of the modifications were ``architectural changes'' with less ``composition'' and ``local refinement'' indicating that larger changes had more impact.
The biggest boost in performance was an ``architectural change'' providing a 4\% boost over its parent in the middle (around days 20 to 30).
As stated earlier, this can be misleading, as at this point multiple well-scoring programs from other island-like groups combined at this point.
Analyzing the parents of this jump, we see three main parent programs injected into the prompt that led to this large jump:

\begin{itemize}
    \item Named ``Resonance Aware Dynamic Arbitrator'', this parent provided the highest performance boost (over 4.5\% IPC boost over the main lineage's parent). This change combined prior ideas on global stride and temporal predictors, checking if predictions matched from the two before issuing a prefetch. If a prefetch was determined to be speculative, then prefetches were redirected to higher cache levels to avoid displacing in-use L1D cache lines.
    \item Named ``Path Aware Arbitrated Prefetcher'', this parent provided a 4.3\% IPC boost over the main lineage's parent.
    This added extra XORing of PC information into predictor table keys to reduce aliasing.
    \item Named ``Refractory Mask Prefetch Filter'', this parent provided a 4\% IPC boost over the main lineage's parent. This added a filter for prefetch requests, temporarily blocking duplicate requests until lower demand accesses or program phase changes.
\end{itemize}

These prior sets of ideas had been developing over time separately from this main lineage in other program database groups but were chosen and combined by the evolutionary database at this time.

\subsection{What were the top ideas?}

To break down monolithic top changes, we further breakdown the entire directed acyclic graph of changes leading to the final evolution output.
This allows understanding of how the monolithic large jump in Figure~\ref{fig:main-lineage} was influenced by what core ideas, as top ideas don't necessarily correspond to the main idea that leads to the change.

Using a custom graph traversal algorithm, we can recursively break down monolithic top ideas into sub-top-ideas.
This is done by recursively looking at lineages starting from the best program.
First, we examine the main lineage and cluster top ideas (relative to their parents).
Of the top clustered ideas, we look at each idea's parent and have an LLM judge determine which parents led to the main impact.
Of those parents, we then treat this as a new lineage, and repeat the clustering loop.

Out of all top ideas that contributed to the final best program, we show the top ideas, relative to the IPC gain over their parent ideas, in Table~\ref{tab:top5}.

\begin{table*}[t!]
\caption{Top ideas breakdown.}
\label{tab:top5}
\centering
\small
\begin{tabularx}{\textwidth}{>{\hsize=0.3\hsize\raggedright\arraybackslash}X l >{\hsize=1.7\hsize\raggedright\arraybackslash}X}

\toprule
\textbf{\% Gain Over Parent} & \textbf{Name} & \textbf{Overview} \\
\midrule
11.0\% & Velocity Aware Delta Engine & Adds ``velocity'' monitoring (measuring instruction retirement) to scale prefetch lookahead distance by adding a new ``global delta table''. \\
\midrule
10.9\% & Execution Velocity Berti Prefetcher & Adds ``velocity'' monitoring (measuring instruction retirement and miss latencies) to scale prefetch lookahead distance by adding a mask per entry. \\
\midrule
7.7\% & Feedback Guided Stride Prefetcher & Replaces old table with new two-level table that better tracks intra- versus inter-page strides, in addition to adapting with bandwidth. \\
\midrule
4.6\% & Cognitive Genome Hybrid & Adds extra XORing of branch paths and PCs to index prefetch tables, ``velocity'' monitoring (instruction retirement), extra masking per prefetch tables (to filter unproductive lookahead), and other smaller changes. \\
\midrule
4.5\% & Context Reward Based Arbitration & Adds an arbitrator between intra/inter-page and stride prefetchers using ``velocity`` monitoring (instruction context, retirement rate, and memory bandwidth). \\
\bottomrule
\end{tabularx}
\end{table*}

Across these ideas, we see five main contributions:

\begin{itemize}
    \item \textit{Execution History and Path-Signature Disambiguation}
    \begin{itemize}
        \item \textit{What:} XORing global branch path history with the PC to index prefetch tables.
        \item \textit{Why:} Helps eliminate instruction aliasing.
    \end{itemize}
    \item \textit{Cache Pollution Mitigation via Depth-Mask Pruning}
    \begin{itemize}
        \item \textit{What:} Adding a bit-vector ``depth mask'' per tracking entry to avoid longer invalid lookahead prefetches.
        \item \textit{Why:} Prevents over-speculation near loop boundaries that causes cache pollution. This pruning tracks historical accuracy per lookahead depth and dynamically truncates deep prefetches, guaranteeing high precision without evicting in-use lines.
    \end{itemize}
    \item \textit{Multi-Level Coordination and Cache Hierarchy Routing}
    \begin{itemize}
        \item \textit{What:} Routes high-confidence prefetches to L1D while diverting speculative or deep-lookahead prefetches to lower-level caches.
        \item \textit{Why:} High-distance prefetches are valuable for hiding DRAM latency but dangerous to small L1D caches thus routing speculative prefetches to L2/LLC warms the lower memory hierarchy ahead of time without displacing critical L1 data.
    \end{itemize}
    \item \textit{Instant Phase-Transition Suppression}
    \begin{itemize}
        \item \textit{What:} Use higher-order derivatives of instruction retirement rates to detect program phase shifts.
        \item \textit{Why:} When code transitions between phases (e.g., from pointer chasing to linear scanning), avoid firing stale prefetches.
    \end{itemize}
    \item \textit{Hardware Realizability and Area/Timing Pruning}
    \begin{itemize}
        \item \textit{What:} Replace monolithic large tables with smaller tables and arbitrate between them.
        \item \textit{Why:} Avoid adding more monolithic tables that might not hit 1-cycle L1 hit timing and area budgets.
    \end{itemize}
\end{itemize}

\begin{figure*}[t]
  \centering
  \includegraphics[width=0.45\linewidth,trim=0 0 0 0,clip]{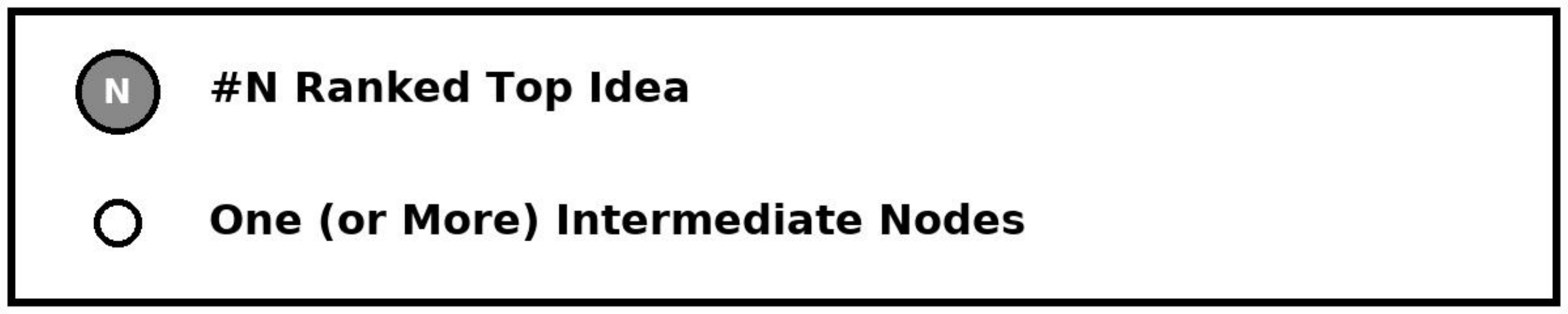}
  \includegraphics[width=1\linewidth,trim=75 75 75 75,clip]{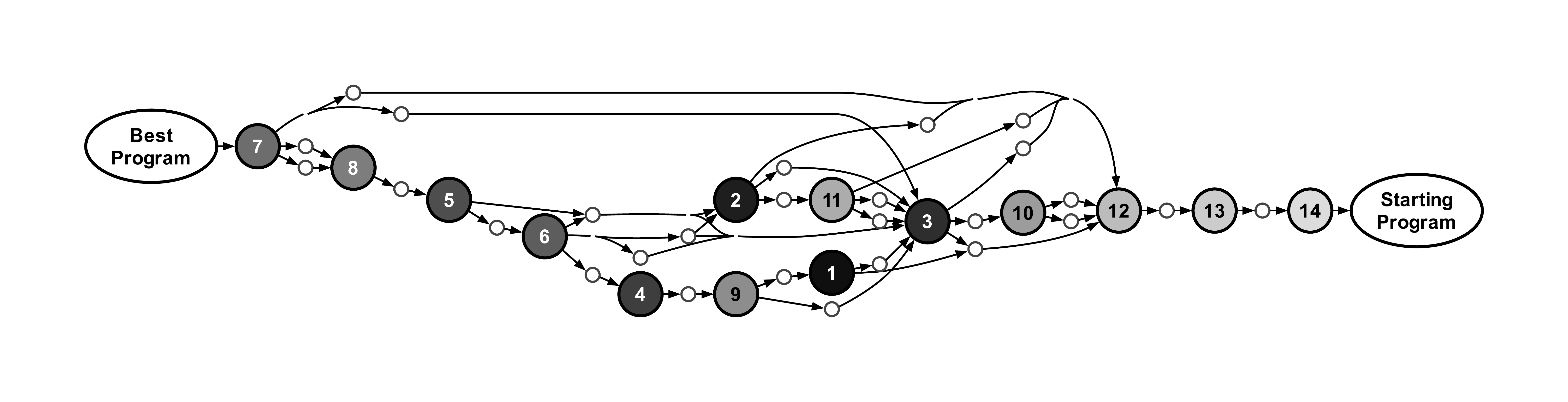}
  \caption{Where ideas are distributed throughout the evolution.}
  \label{fig:ideas-from-where}
\end{figure*}

Figure~\ref{fig:ideas-from-where} visualizes where top ideas are located through the evolutionary run represented as a directed acyclic graph.
Numbered nodes represent top ideas numbered based on how much improvement was given over their parents (e.g., \#3 is the 3rd best top idea).
Small white nodes represent one or more intermediate nodes in between best ideas that bridge different branches/lineages together. 

Here we see that sub-top-ideas come from throughout the evolution on many different branches.
Some of the top best ideas (i.e., \#2 and \#3) are seeds for multiple branches, indicating their importance and widespread effect across all branches.

\section{Related Work and Discussions}

This work is complementary to the work that is happening to improve both existing agentic harnesses for optimization as well as improvements in LLMs.
We expect improvements in both spaces to yield larger gains more rapidly and with better efficiency as more reasoning and context is provided to newer LLMs.
For this work, we focus on using \anonevolve~\cite{novikov2025alphaevolve} since this work is an extension of \alphaarchitect's original framework.

The usage of such agentic tooling in computer architecture domains is growing in popularity.
Since the November 2025 publication of \alphaarchitect~\cite{Venues}, more work in both optimization~\cite{blasberg2026agentic} and full-scale designs~\cite{sankaralingam2026computerarchitecturesalphazeromoment, cui2026chiaopensourceframeworkprincipled} has occurred across domains in computer architecture.
We expect more such agentic work to grow in the future with the proliferation of more dedicated workshops and conferences in the space~\cite{mlarchsys, iclad} and improvements in LLMs/harnesses.
This work builds on the original \alphaarchitect line of work, but has general insights for any harness built around computer architecture in the future.

For prefetcher design and optimization work, there is a long history of human-created prefetchers.
Instead of focusing on any particular generated design, this work emphasizes that \alphaarchitect can generate competitive prefetcher designs automatically, demonstrating one such example.
While the prefetcher generated is able to beat the prior SoTA prefetcher design, more work is needed to have the tool emit significant novel advancements.
First, in the L1D prefetcher generated, its RL-based arbitration (contrasting with standalone RL prefetchers~\cite{bera2021pythia,10.1145/2749469.2749473}), GST's changing stride tracker (contrasting with constant stride tracking), and extra long-lived tracking metadata added (contrasting with prior work using rewards when data arrives~\cite{10.1145/3725843.3756096, siddiqui2025coordinatedreinforcementlearningprefetching}) provide limited novelty.
Secondly, in the LLC prefetcher design, the arbitrator is incrementally novel, since it is able to count how many unique steps occurred in a time period for more accurate arbitration, compared to prior works that use simple hit/miss thresholds~\cite{10.1145/2540708.2540730}.
These changes generally emphasize the strengths of LLM tooling as good synthesizers, combining multiple domains together, while providing limited novelty.

\section{Conclusion}

In this paper we have extended \alphaarchitect v1 towards building and optimizing data prefetching algorithms in the context of the Fourth Data Prefetching Competition.
Through both providing more hardware realizability area feedback and decomposing the problem into a divide-and-conquer cascade, we show that this \alphaarchitect v2 methodology is able to beat human-derived, winning policies in the competition setting.
We show that this methodology is able to provide a 3.8\% increase in performance versus the competition baseline and beat the existing state-of-the-art winner, BertiGO, by 0.3\%.
Additionally, we analyze over 12,000 ideas from one of the cascaded evolutions, describing the breadth of top ideas and why they were impactful, giving a glimpse into how these agentic evolutionary discovery tools explore the vast design space.
As AI, LLMs, and their harnesses for scientific discovery improve, this work demonstrates one example of AI-augmented computer architecture discovery.


\bibliographystyle{IEEEtranS}
\bibliography{refs}

@article{novikov2025alphaevolve,
      title={AlphaEvolve: A coding agent for scientific and algorithmic discovery}, 
      author={Alexander Novikov and Ngân Vũ and Marvin Eisenberger and Emilien Dupont and Po-Sen Huang and Adam Zsolt Wagner and Sergey Shirobokov and Borislav Kozlovskii and Francisco J. R. Ruiz and Abbas Mehrabian and M. Pawan Kumar and Abigail See and Swarat Chaudhuri and George Holland and Alex Davies and Sebastian Nowozin and Pushmeet Kohli and Matej Balog},
      year={2025},
      eprint={2506.13131},
      archivePrefix={arXiv},
      primaryClass={cs.AI},
      journal={arXiv preprint arXiv:2506.13131},
      url={https://arxiv.org/abs/2506.13131}, 
}

@misc{openevolve,
  title = {OpenEvolve: an open-source evolutionary coding agent},
  author = {Asankhaya Sharma},
  year = {2025},
  publisher = {GitHub},
  url = {https://github.com/algorithmicsuperintelligence/openevolve}
}

@book{tanese1989distributed,
  title={Distributed genetic algorithms for function optimization},
  author={Tanese, Reiko},
  year={1989},
  publisher={University of Michigan}
}

@article{cheng2025barbarians,
      title={Barbarians at the Gate: How AI is Upending Systems Research}, 
      author={Audrey Cheng and Shu Liu and Melissa Pan and Zhifei Li and Bowen Wang and Alex Krentsel and Tian Xia and Mert Cemri and Jongseok Park and Shuo Yang and Jeff Chen and Lakshya Agrawal and Aditya Desai and Jiarong Xing and Koushik Sen and Matei Zaharia and Ion Stoica},
      year={2025},
      journal={arXiv preprint arXiv:2510.06189},
      eprint={2510.06189},
      archivePrefix={arXiv},
      primaryClass={cs.AI},
      url={https://arxiv.org/abs/2510.06189}, 
}

@inproceedings{liu2026skydiscover,
  title={SkyDiscover: A Flexible, Adaptive Framework for AI-Driven Scientific and Algorithmic Discovery},
  author={Liu, Shu and Cemri, Mert and Agarwal, Shubham and Krentsel, Alexander and Naren, Ashwin and Mang, Qiuyang and Li, Zhifei and Gupta, Akshat and Maheswaran, Monishwaran and Cheng, Audrey and others},
  booktitle={Proceedings of the ACM Conference on AI and Agentic Systems},
  pages={1223--1227},
  year={2026}
}

@misc{sankaralingam2026computerarchitecturesalphazeromoment,
      title={Computer Architecture's AlphaZero Moment: Automated Discovery in an Encircled World}, 
      author={Karthikeyan Sankaralingam},
      year={2026},
      eprint={2604.03312},
      archivePrefix={arXiv},
      primaryClass={cs.AR},
      url={https://arxiv.org/abs/2604.03312}, 
}

@article{gober2022championship,
  title={The championship simulator: Architectural simulation for education and competition},
  author={Gober, Nathan and Chacon, Gino and Wang, Lei and Gratz, Paul V and Jimenez, Daniel A and Teran, Elvira and Pugsley, Seth and Kim, Jinchun},
  journal={arXiv preprint arXiv:2210.14324},
  year={2022}
}

@misc{hamadanian2026gliahumaninspiredaiautomated,
      title={Glia: A Human-Inspired AI for Automated Systems Design and Optimization}, 
      author={Pouya Hamadanian and Pantea Karimi and Arash Nasr-Esfahany and Kimia Noorbakhsh and Joseph Chandler and Ali ParandehGheibi and Mohammad Alizadeh and Hari Balakrishnan},
      year={2026},
      eprint={2510.27176},
      archivePrefix={arXiv},
      primaryClass={cs.AI},
      url={https://arxiv.org/abs/2510.27176}, 
}

@misc{Google_Workload_Traces_Version_2,
  author={Google},
  title = {Google Workload Traces Version 2},
  howpublished = {\url{https://console.cloud.google.com/storage/browser/external-traces-v2}},
  note = {Accessed: 2025-11-13}
}

@misc{pelleriti2026evolutionarycodingagentsevolve,
      title={What Do Evolutionary Coding Agents Evolve?}, 
      author={Nico Pelleriti and Sree Harsha Nelaturu and Zhanke Zhou and Zongze Li and Max Zimmer and Bo Han and Sebastian Pokutta},
      year={2026},
      eprint={2605.20086},
      archivePrefix={arXiv},
      primaryClass={cs.NE},
      url={https://arxiv.org/abs/2605.20086}, 
}

@article{cemri2026multi,
  title={Why do multi-agent llm systems fail?},
  author={Cemri, Mert and Pan, Melissa Z and Yang, Shuyi and Agrawal, Lakshya A and Chopra, Bhavya and Tiwari, Rishabh and Keutzer, Kurt and Parameswaran, Aditya and Klein, Dan and Ramchandran, Kannan and others},
  journal={Advances in Neural Information Processing Systems},
  volume={38},
  year={2026}
}

@article{gupta2026archagent,
  title={ArchAgent: Agentic AI-driven Computer Architecture Discovery},
  author={Gupta, Raghav and Jain, Akanksha and Gonzalez, Abraham and Novikov, Alexander and Huang, Po-Sen and Balog, Matej and Eisenberger, Marvin and Shirobokov, Sergey and V{\~u}, Ng{\^a}n and Dixon, Martin and others},
  journal={arXiv preprint arXiv:2602.22425},
  year={2026}
}

@article{singh2026pushing,
  title={Pushing the Limits of the Berti Prefetcher},
  author={Singh, Simranjit and Torres, AN and Ros, Alberto},
  journal={4th Data Prefetching Championship (DPC4)},
  year={2026}
}

@article{journals/nature/RomeraParedesBNBKDREWFKF24,
  author = {Romera-Paredes, Bernardino and Barekatain, Mohammadamin and Novikov, Alexander and Balog, Matej and Kumar, M. Pawan and Dupont, Emilien and Ruiz, Francisco J. R. and Ellenberg, Jordan S. and Wang, Pengming and Fawzi, Omar and Kohli, Pushmeet and Fawzi, Alhussein},
  ee = {https://doi.org/10.1038/s41586-023-06924-6},
  journal = {Nat.},
  month = {January},
  number = 7995,
  pages = {468-475},
  title = {Mathematical discoveries from program search with large language models.},
  url = {http://dblp.uni-trier.de/db/journals/nature/nature625.html#RomeraParedesBNBKDREWFKF24},
  volume = 625,
  year = 2024
}

@misc{mouret2015illuminatingsearchspacesmapping,
      title={Illuminating search spaces by mapping elites}, 
      author={Jean-Baptiste Mouret and Jeff Clune},
      year={2015},
      eprint={1504.04909},
      archivePrefix={arXiv},
      primaryClass={cs.AI},
      url={https://arxiv.org/abs/1504.04909}, 
}

@article{posluns2026global,
  title={Global Berti: Simultaneous Streaming and Spatial Prefetching},
  author={Posluns, Gilead and Jeffrey, Mark C},
  year = {2026},
  month = feb,
  journal = {4th Data Prefetching Championship}
}

@inproceedings{10.1145/2749469.2749473,
author = {Peled, Leeor and Mannor, Shie and Weiser, Uri and Etsion, Yoav},
title = {Semantic locality and context-based prefetching using reinforcement learning},
year = {2015},
isbn = {9781450334020},
publisher = {Association for Computing Machinery},
address = {New York, NY, USA},
url = {https://doi.org/10.1145/2749469.2749473},
doi = {10.1145/2749469.2749473},
booktitle = {Proceedings of the 42nd Annual International Symposium on Computer Architecture},
pages = {285–297},
numpages = {13},
location = {Portland, Oregon},
series = {ISCA '15}
}

@inproceedings{10.1145/3725843.3756096,
author = {Block, Charles and Gerogiannis, Gerasimos and Torrellas, Josep},
title = {Micro-MAMA: Multi-Agent Reinforcement Learning for Multicore Prefetching},
year = {2025},
isbn = {9798400715730},
publisher = {Association for Computing Machinery},
address = {New York, NY, USA},
url = {https://doi.org/10.1145/3725843.3756096},
doi = {10.1145/3725843.3756096},
booktitle = {Proceedings of the 58th IEEE/ACM International Symposium on Microarchitecture},
pages = {884–898},
numpages = {15},
location = {
},
series = {MICRO '25}
}

@misc{Codex, author={OpenAI}, url={https://chatgpt.com/codex/}, journal={Codex in CHATGPT | AI coding agents for Software Engineering}}

@misc{Claude, author={Anthropic}, url={https://claude.com/product/claude-code}, journal={Claude}}

@misc{GoogleAntigravity, author={Google}, url={https://antigravity.google/}, journal={Google Antigravity}}

@misc{opencode2026,
  author = {{SST} and {OpenCode Contributors}},
  title = {OpenCode: The Open-Source AI Coding Agent},
  year = {2026},
  url = {https://github.com/anomalyco/opencode},
  note = {Accessed: 2026-07-31}
}

@misc{karimi2026improvingcoherencepersistenceagentic,
      title={Improving Coherence and Persistence in Agentic AI for System Optimization}, 
      author={Pantea Karimi and Kimia Noorbakhsh and Mohammad Alizadeh and Hari Balakrishnan},
      year={2026},
      eprint={2603.21321},
      archivePrefix={arXiv},
      primaryClass={cs.AI},
      url={https://arxiv.org/abs/2603.21321}, 
}

@inproceedings{10.1145/2540708.2540730,
author = {Jain, Akanksha and Lin, Calvin},
title = {Linearizing irregular memory accesses for improved correlated prefetching},
year = {2013},
isbn = {9781450326384},
publisher = {Association for Computing Machinery},
address = {New York, NY, USA},
url = {https://doi.org/10.1145/2540708.2540730},
doi = {10.1145/2540708.2540730},
booktitle = {Proceedings of the 46th Annual IEEE/ACM International Symposium on Microarchitecture},
pages = {247–259},
numpages = {13},
location = {Davis, California},
series = {MICRO-46}
}

@misc{siddiqui2025coordinatedreinforcementlearningprefetching,
      title={Coordinated Reinforcement Learning Prefetching Architecture for Multicore Systems}, 
      author={Mohammed Humaid Siddiqui and Fernando Guzman and Yufei Wu and Ruishu Ann},
      year={2025},
      eprint={2509.10719},
      archivePrefix={arXiv},
      primaryClass={cs.DC},
      url={https://arxiv.org/abs/2509.10719}, 
}

@article{navarroentangling,
  title={The Entangling Data Prefetcher},
  author={Navarro-Torres, Agust{\'\i}n and Singh, Simranjit and Panda, Biswabandan and Ros, Alberto},
  year = {2026},
  month = feb,
  journal = {4th Data Prefetching Championship}
}

@article{meng2026scientistone,
  title     = {ScientistOne: Towards Human-Level Autonomous Research via Chain-of-Evidence},
  author    = {Meng, Rui and Dalvi Mishra, Bhavana and Chen, Jiefeng and Li, Chun-Liang and Goyal, Palash and Parmar, Mihir and Song, Yiwen and Song, Yale and Sinha, Rajarishi and Ranganathan, Parthasarathy and Gokturk, Burak and Yoon, Jinsung and Pfister, Tomas},
  journal   = {arXiv preprint},
  year      = {2026}
}

@misc{Venues,
  title        = {{ArchAgent}: Agentic {AI}-driven Computer Architecture Discovery},
  author       = {Anonymous},
  year         = {2025},
  howpublished = {OpenReview Anonymous Preprint},
  note         = {Submission Number 714},
  url          = {https://openreview.net/forum?id=hcxN9l6zqZ}
}

@article{blasberg2026agentic,
  title={Agentic Architect: An Agentic AI Framework for Architecture Design Exploration and Optimization},
  author={Blasberg, Alexander and Kypriotis, Vasilis and Skarlatos, Dimitrios},
  journal={arXiv preprint arXiv:2604.25083},
  year={2026}
}

@inproceedings{navarro2022berti,
  title={Berti: an accurate local-delta data prefetcher},
  author={Navarro-Torres, Agust{\'\i}n and Panda, Biswabandan and Alastruey-Bened{\'e}, Jes{\'u}s and Ib{\'a}{\~n}ez, Pablo and Vi{\~n}als-Y{\'u}fera, V{\'\i}ctor and Ros, Alberto},
  booktitle={2022 55th IEEE/ACM International Symposium on Microarchitecture (MICRO)},
  pages={975--991},
  year={2022},
  organization={IEEE}
}

@misc{cui2026chiaopensourceframeworkprincipled,
      title={CHIA: An open-source framework for principled, agentic AI-driven hardware/software co-design research}, 
      author={Angela Cui and Ferran Hermida-Rivera and Jack Toubes and Raghav Gupta and Jim Fang and Chengyi Lux Zhang and Ella Schwarz and Junha Kim and Yakun Sophia Shao and Borivoje Nikolic and Christopher W. Fletcher and Sagar Karandikar},
      year={2026},
      eprint={2606.27350},
      archivePrefix={arXiv},
      primaryClass={cs.AR},
      url={https://arxiv.org/abs/2606.27350}, 
}

@inproceedings{bera2021pythia,
  title={Pythia: A customizable hardware prefetching framework using online reinforcement learning},
  author={Bera, Rahul and Kanellopoulos, Konstantinos and Nori, Anant and Shahroodi, Taha and Subramoney, Sreenivas and Mutlu, Onur},
  booktitle={MICRO-54: 54th Annual IEEE/ACM International Symposium on Microarchitecture},
  pages={1121--1137},
  year={2021}
}

@misc{Home, title={Home}, author={Bera, Rahul and Kanellopoulos, Konstantinos and Mutlu, Onur}, url={https://sites.google.com/corp/view/dpc4-2026/home?authuser=0}, journal={DPC4: The 4th Data Prefetching Championship}}

@misc{mlarchsys, title={MLArchSys 2026 Home}, author={Asgari, Bahar and Doudali, Thaleia D. and Huang, Qijing and Jain, Akanksha and Jeong, Geonhwa and St. John, Tom and Panda, Priya and Pandey, Santosh and Subramanian, Suvinay and Yadwadkar, Neeraja J. and Yazdanbakhsh, Amir}, url={https://sites.google.com/corp/view/mlarchsys}, journal={ML for Computer Architecture and Systems}}

@misc{iclad, title={IEEE International Conference on LLM-Aided Design, 2026}, author={Mirhosseini, Azaliza and Liu, Yong}, url={https://iclad.ai/}, journal={IEEE International Conference on LLM-Aided Design}}

@misc{Hparch, url={https://comparch-conf.gatech.edu/dpc2/}, journal={DPC-CFP}, author={Hparch}}

@misc{Pico, author={Alameldeen, Alaa and Pugsley, Seth}, url={https://dpc3.compas.cs.stonybrook.edu/}, journal={Pico}}

@misc{alameldeenfirst,
  title={The First JILP Data Prefetching Championship (DPC-1)},
  author={Alameldeen, Alaa R. and Rotenberg, Eric},
  url={https://jilp.org/dpc/}
}

@article{lange2025shinkaevolve,
  title={Shinkaevolve: Towards open-ended and sample-efficient program evolution},
  author={Lange, Robert Tjarko and Imajuku, Yuki and Cetin, Edoardo},
  journal={arXiv preprint arXiv:2509.19349},
  year={2025}
}

\end{document}